%% file: main.tex
\documentclass[10pt,twocolumn,letterpaper]{article}
\usepackage{cvpr}
\usepackage[utf8]{inputenc}
\usepackage[T1]{fontenc}
\usepackage{amsfonts}
\usepackage{amsmath}
\usepackage{booktabs}
\usepackage{graphicx}
\usepackage{microtype}
\usepackage{multicol}
\usepackage{nicefrac}
\usepackage{soul}
\usepackage{url}
\usepackage[normalem]{ulem}
\usepackage{xcolor}
\usepackage{amssymb}
\usepackage{pdfrender}
\usepackage{pifont}
\usepackage[pagebackref=true,breaklinks=true,letterpaper=true,colorlinks,bookmarks=false,citecolor=orange]{hyperref}
\usepackage{cleveref}
\usepackage{circledsteps}
\usepackage{cuted}

\useunder{\uline}{\ul}{}
\definecolor{maroon}{HTML}{efa884}
\definecolor{dorange}{HTML}{3b2000}
\hypersetup{citecolor=maroon}
\hypersetup{linkcolor=maroon}
\hypersetup{linkcolor=maroon}
\hypersetup{urlcolor=orange}

\input{macros}

\makeatletter
\renewcommand{\paragraph}{%
  \@startsection{paragraph}{4}%
  {\z@}{0.25em}{-1em}%
  {\normalfont\normalsize\bfseries}%
}
\makeatother

\def\cvprPaperID{*****}
\def\confName{CVPR}
\def\confYear{2023}

\begin{document}

\title{Common Pets in 3D\@:\\
Dynamic New-View Synthesis of Real-Life Deformable Categories}

\author{Samarth Sinha\\
University of Toronto\\
{\tt\small samarth.sinha@mail.utoronto.ca}
\and
Roman Shapovalov \\
Meta AI\\
{\tt\small romansh@meta.com}
\and
Jeremy Reizenstein\\
Meta AI\\
{\tt\small reizenstein@meta.com}
\and
Ignacio Rocco\\
Meta AI\\
{\tt\small irocco@meta.com}
\and
Natalia Neverova\\
Meta AI \\
{\tt\small nneverova@meta.com}
\and
Andrea Vedaldi\\
Meta AI\\
{\tt\small vedaldi@meta.com}
\and
David Novotny\\
Meta AI\\
{\tt\small dnovotny@meta.com}
}
\twocolumn[{%
\renewcommand\twocolumn[1][]{#1}%
\maketitle
\thispagestyle{empty}
\vspace{-1.4cm}
\begin{center}
\centering
\captionsetup{type=figure}
\includegraphics[width=\textwidth]{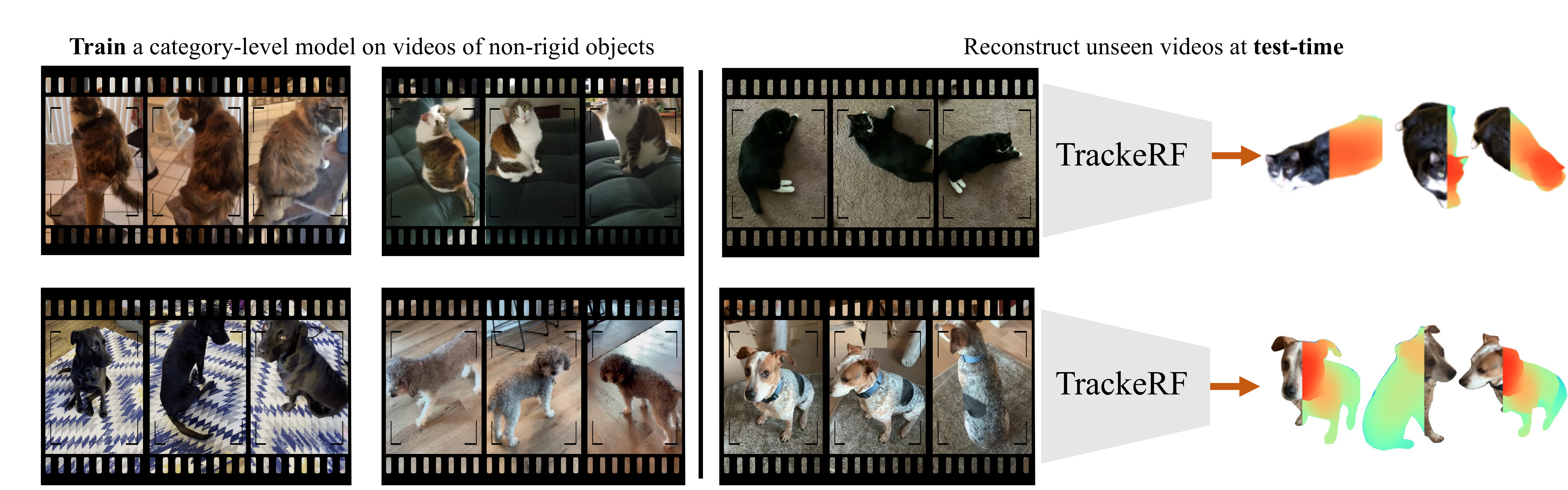}
\vspace{-0.7cm}
\captionof{figure}{
We tackle the problem of synthesising new views of deformable objects given only a small number of views taken at different times.
We introduce new benchmark data for this task: \textbf{\thedataset} (\thedatasetabbrev), containing 4,200 smartphone videos of cats and dogs collected `in the wild'.
We also propose a new method, \textbf{\themethod}, a deformable new-view synthesis algorithm which learns a category-level reconstruction prior from videos and applies it to reconstruct new objects at test time.}%
\label{fig:teaser-fig}%
\end{center}%
\vspace{0.5em}
}]

\begin{abstract}
Obtaining photorealistic reconstructions of objects from sparse views is inherently ambiguous and can only be achieved by learning suitable reconstruction priors. 
Earlier works on sparse rigid object reconstruction successfully learned such priors from large datasets such as CO3D.
In this paper, we extend this approach to dynamic objects.
We use cats and dogs as a representative example and introduce \emph{\thedataset} (\thedatasetabbrev), a collection of crowd-sourced videos showing around 4,200 distinct pets.
\thedatasetabbrev is one of the first large-scale datasets for benchmarking non-rigid 3D reconstruction ``in the wild''.
We also propose \themethod, a method for learning 4D reconstruction from our dataset.
At test time, given a small number of video frames of an unseen object, \themethod predicts the \textit{trajectories} of its 3D points and generates new views, interpolating viewpoint and time.
Results on \thedatasetabbrev reveal significantly better non-rigid new-view synthesis performance than existing baselines.
\end{abstract}

\input{001_intro}

\input{002_rw}

\input{003_cp3d}
\input{004_background}
\input{005_method}

\input{006_exps}

\input{007_conclusion}

{\small\bibliographystyle{splncs04}\bibliography{refs,vedaldi_general}}

\clearpage\appendix
\input{008_appendix}

\end{document}

%% file: macros.tex
\makeatletter
\newcommand{\oneptsmaller}[1]{%
  \begingroup
  \fontsize{\dimexpr\f@size pt-1pt}{\f@baselineskip}\selectfont
  #1%
  \endgroup
}
\makeatother

\newcommand{\thedataset}{Common Pets in 3D\xspace}
\newcommand{\thedatasetabbrev}{CoP3D\xspace}
\newcommand{\themethod}{Tracker-NeRF\xspace}

\definecolor{trackerfcolor}{HTML}{361e08}

\newcommand{\themethodabbrev}{\textcolor{trackerfcolor}{TrackeRF}\xspace}

\newcommand{\setfont}{\ttfamily}

\newcommand{\traintestset}{{\setfont train-unseen}\xspace}

\newcommand{\testtestset}{{\setfont test-unseen}\xspace}

\newcommand{\bx}{\mathbf{x}}
\newcommand{\bc}{\mathbf{c}}
\newcommand{\br}{\mathbf{r}}
\newcommand{\bz}{\mathbf{z}}
\newcommand{\bu}{\mathbf{u}}
\newcommand{\bC}{\mathbf{C}}

\newcommand{\method}[1]{{\oneptsmaller{\sffamily #1}}}
\newcommand{\tgt}{\text{tgt}}
\newcommand{\src}{\text{src}}
\newcommand{\tsrc}{t^\src}

\newcommand{\Itgt}{I^\text{tgt}}
\newcommand{\barItgt}{{\bar I}^\text{tgt}}
\newcommand{\Mtgt}{M^\text{tgt}}
\newcommand{\barMtgt}{{\bar M}^\text{tgt}}
\newcommand{\Isrc}{I^\text{src}}
\newcommand{\Ptgt}{P^\text{tgt}}
\newcommand{\Psrc}{P^\text{src}}

\newcommand{\xmark}{\scalebox{0.85}{\ding{53}}}%

\definecolor{bulgarianrose}{rgb}{0.28, 0.02, 0.03}
\definecolor{forestgreen}{rgb}{0.0, 0.27, 0.13}

\newcommand*{\boldcheckmark}{%
  \textpdfrender{
    TextRenderingMode=FillStroke,
    LineWidth=.5pt, 
  }{\textcolor{forestgreen}\checkmark}%
}

\newcommand*{\boldxmark}{%
  \textpdfrender{
    TextRenderingMode=FillStroke,
    LineWidth=.5pt, 
  }{\textcolor{bulgarianrose}\xmark}%
}

%% file: 001_intro.tex
\section{Introduction}%
\label{sec:intro}

Advances in photorealistic reconstruction and new-view synthesis facilitate experiencing real-life objects and scenes in virtual and mixed reality.
However, compared to established technologies, such as digital photography, capturing 3D content remains significantly more difficult and limited.
Methods based on neural radiance fields~\cite{Lombardi2019nv,Mildenhall2020,Liu2020nsvf}, signed distance functions~\cite{yariv20multiview}, and other implicit representations~\cite{sitzmann19scene} use deep neural networks to represent the geometry of a single scene~\cite{Mildenhall2020}, and generally require hundreds of input views for high-quality reconstruction.
Furthermore, they are limited to the reconstruction of rigid objects and static scenes.
Instead, users should be able to capture 3D and 4D content as easily as taking an image or video with their smartphone, a setting that we call \emph{casual capture}.

Casual capture requires reconstructing objects from only a few views, which is inherently ambiguous.
This is particularly true for dynamic content because it requires to reconstruct not only shape and appearance, but also deformations.
Deformations can be controlled via regularization~\cite{park20deformable,Tretschk2020}, but such constraints may be too weak or too restrictive to work well in all cases.
Alternatively, one can use parametric models of the objects, such as skeletons or skinned surface models~\cite{peng20neural,Liu2021neuralactor,Weng2022humannerf,anerf,neuman,vid2actor} for humans, but their design is a major engineering challenge that cannot easily scale to arbitrary content.

Methods such as~\cite{reizenstein21common,chen21snarf:,codenerf,autorf,yu20pixelnerf:,wce} address the issue of ambiguity by learning \emph{3D priors} for specific types of objects from large datasets such as CO3D~\cite{co3d}.
With these priors, they can achieve plausible 3D reconstructions from a small number of views, but so far only for rigid objects.
In this paper, we wish to further extend this data-driven approach to \emph{dynamic objects} which change their shape \emph{during capture} and thus require a 4D reconstruction.
Our hypothesis is that dynamic objects can be reconstructed even from a monocular video, provided that one leverages priors learnt from a large collection of related videos.

In order to test our hypothesis, we first contribute a new crowd-sourced dataset, \emph{Common Pets in 3D} (\thedatasetabbrev).
\thedatasetabbrev contains 4,200 videos of distinct cats and dogs.
It contains more than 600,000 video frames with 3D camera tracking obtained using Structure-from-Motion (SfM) and foreground object masks for each frame.
The videos are captured \textit{in the wild} by non-experts, using smartphone cameras, and are thus representative of the casual capture setting.
Like CO3D, \thedatasetabbrev can be used for single-sequence and category-wise reconstruction.

Our second contribution is \emph{\themethod} (\themethodabbrev), a new method for few-view reconstruction of dynamic objects.
The method is inspired by approaches such as Warped Ray Embedding~\cite{henzler2021unsupervised}, PixelNeRF~\cite{yu20pixelnerf:}, and NeRFormer~\cite{reizenstein21common} in that it learns to synthesise new views by triangulating features extracted form the provided input views.
Our key innovation is modelling the deformations of objects by inferring the \emph{3D trajectory} of each queried 3D point.
Empirically, we show that \themethodabbrev outperforms previous approaches for new-view synthesis of dynamic objects.

%% file: 002_rw.tex
\section{Related work}%
\label{sec:rw}

\begin{figure*}[t!]
\centering
\includegraphics[width=\textwidth]{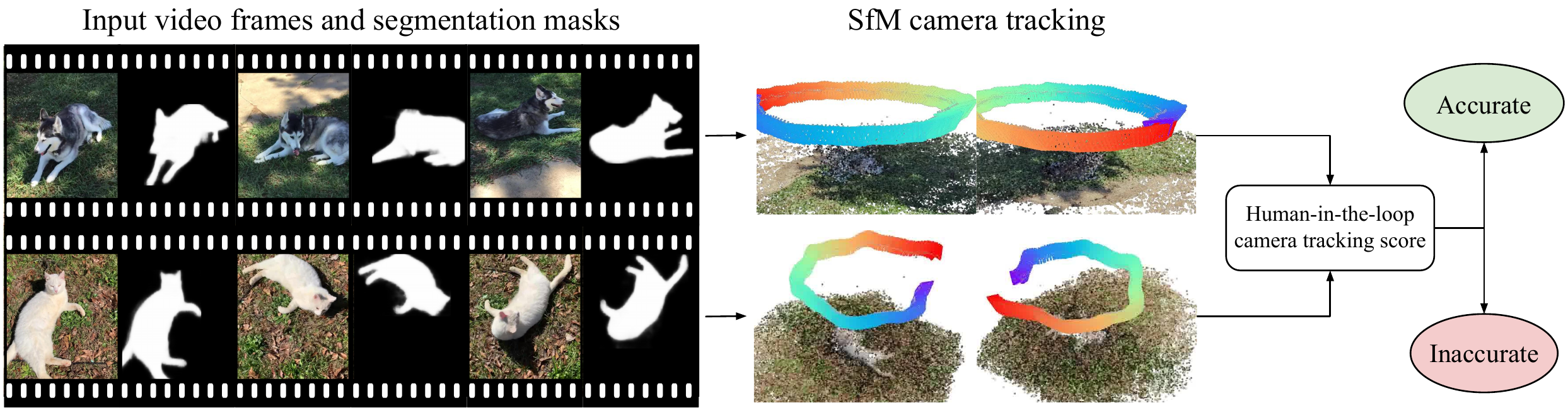}
\caption{
\textbf{\thedataset} (\thedatasetabbrev) comprises 360$^{\circ}$ videos of pets collected using Amazon Mechanical Turk.
The camera in each video is tracked using SfM (COLMAP~\cite{schoenberger2016sfm}).
The quality of the camera tracking is assessed using active learning.}%
\label{fig:cp3d}
\end{figure*}

\paragraph{Dynamic new-view synthesis.}

Various scene representations have been explored for new-view synthesis, including meshes~\cite{UMR,Tulsiani2020}, voxels~\cite{rematas21sharf:,plenoxels,kar2017learning,neuralvolumes}, radiance fields~\cite{Mildenhall2020,ngp,blocknerf}, and sign-distance functions~\cite{sitzmann19scene,niemeyer19occupancy,neuralparts}.
Most of these works have considered static scenes, but a few have explored extensions to dynamic scenes.

The simplest extension is to add time as an additional parameter of the representation~\cite{Li2021nsff,videonerf}.
For example, NSFF~\cite{Li2021nsff} adds time to NeRF, defining a space-time radiance field.
The latter is regularized by fitting a scene flow field, using it to enforce time consistency.
Video-NeRF~\cite{videonerf} fits a space-time radiance field too, but regularizes it by using a monocular depth predictor.

An alternative approach is to use a deformation field to reduce the reconstruction of the different video frames to a canonical representation which is exactly or approximately time-invariant~\cite{Lombardi2019nv,nerfflow,Pumarola2020,nrnerf}.
Neural Volumes~\cite{Lombardi2019nv} learns an auto-encoder model mapping several input frames to an instantaneous low-dimensional latent code; the latter is decoded into a radiance field which is approximately time invariant and a corresponding deformation field, expressed as a mixture of a small number of local affine transformations.
Closer to NeRF, D-Nerf~\cite{Pumarola2020} does not use an auto-encoder, but fits a canonical neural radiance field directly to a single video together with a corresponding deformation field.
Nerfies~\cite{park20deformable} fits an auto-decoder instead, learning small codes parameterising the appearance and deformation of the individual video frames.
The deformation is a field of $SE(3)$ transformations whose complexity is controlled by an elastic regularization term.
Non-rigid NeRF~\cite{Tretschk2020} also uses an auto-decoder, but only to express deformations, which controls the magnitude, and discourages stretching and squeezing.

Other methods regularize the scene flow by exploiting pre-trained optical flow predictors such as RAFT~\cite{teed20raft:}.
For instance,
LASR~\cite{yang21lasr:} fits a mesh predictor to a single video of a deformable object and supervises it by using a combination of photometric and optical flow losses.

\paragraph{Category-specific new-view synthesis.}

Like NeRF, most of the methods above start from a `tabula rasa', fitting a model to a single video sequence.
However, many reconstruction tasks are inherently ambiguous and can be approached successfully only by combining evidence with prior information.
One way to do so is to learn an auto-encoder from many videos of a given object category.
Most such methods work by fitting a parametric model of an articulated object.
CMR~\cite{kanazawa18learning} and~\cite{kokkinos2021learning} rely on sparse keypoint supervision to deform a template mesh to fit a given image.
UMR~\cite{li20self-supervised} does not require keypoints but assumes part segmentation learned in a self-supervised manner.
DOVE~\cite{wu21dove:} trains an auto-encoder from many videos of an object category.
The decoder is fixed and given by a deformable texture mesh model of the object.
Optical flow is used as additional supervision.
BANMo~\cite{Yang2021banmo} fits a volumetric model to multiple videos of a single object instance, warping it using neural blend skinning.
This method also uses category-level priors in the form of pre-trained canonical surface embeddings~\cite{Neverova2020cse}.

\paragraph{Sparse new-view synthesis.}

The methods above require many input images for each reconstructed object.
Reconstruction from few views, or even a single view, further emphasises the importance of priors.
PixelNeRF~\cite{Yu2020pixelnerf} and WCE~\cite{henzler2021unsupervised} learn a category-specific multi-view auto-encoder for a radiance field.
In order to predict colour and occupancy of a 3D point, they reproject it on the available views and pool the corresponding 2D features.
NeRFormer~\cite{reizenstein21common} and ViewFormer~\cite{viewformer} generalise these architectures by using a transformer to pool information across views and along rays~\cite{vaswani17attention}.
However, these methods assume that the observed object is rigid, or are limited to using a single view.
We contribute instead the first dynamic new-view synthesis method that can use a sparse set of views as input.

\paragraph{Datasets for new-view synthesis.}

There are only a few real-life datasets that can be used to train models for category-specific new-view synthesis.
Choi et al.~\cite{Choi2016tanks} and GSO~\cite{ignition2020google} introduced datasets of 2,000 real-world videos each.
Each video shows around the object and has ground-truth depth.
Objectron~\cite{ahmadyan2020objectron} introduced 15,000 videos, but only some of them show the object from all sides.
Dove~\cite{wu21dove:} provided a dataset consisting of a few long videos of birds shot from static cameras~\cite{wu21dove:}, but with comparatively little variety.
The CO3D dataset~\cite{reizenstein21common} contains 19,000 fly-around videos of rigid objects of 50 common categories.
Like CO3D, our new dataset provides fly-around videos of the objects, with the difference that objects deform over time (they are pets).
The latter makes our \thedatasetabbrev a perfect candidate for studying new-view synthesis of deformable objects.

%% file: 003_cp3d.tex
\section{\thedataset: a new dataset}

Our new \emph{Common Pets in 3D} (\thedatasetabbrev) dataset contains 4,200 videos of cats and dogs of different breeds, captured under different viewpoint, camera motion, background and illumination, and moving in different ways.
Videos show around the pets and are collected `in the wild' via crowdsourcing.
Compared to CO3D~\cite{co3d}, which is collected in a similar manner, reconstruction is much more challenging because pets can move over time, sometimes suddenly.

\thedatasetabbrev is designed as a benchmark for new-view synthesis.
Hence, in addition to the videos, \thedatasetabbrev contains intrinsic and extrinsic camera parameters and masks of the reconstructed objects.
It is a collection of videos
$
\mathcal{V}_i = \{ (I_i^j,M_i^j,P_i^j,t_i^j) \}_{j=1}^{N_{\mathcal{V}_i}},
$
each consisting of a sequence of $N_{\mathcal{V}_i}$ RGB
frames $I_i^j\in\mathbb{R}^{3 \times H \times W}$, masks $M_i^j \in [0, 1]^{H \times W}$, cameras  $P_i^j \in \mathbb{R}^{4 \times 4}$ and time stamps $t_i^j \in \mathbb{R}_+$.

\paragraph{Data collection.}

Videos in \thedatasetabbrev were collected by asking Amazon Mechanical Turk (AMT) workers to capture 360$^\circ$ videos of their pets using a smartphone.
Each video was manually reviewed for quality and to ensure that pet owners acted responsibly, ensuring the safety of their pets, others, and themselves.
The videos focus on pets and do not contain personal information such as human faces.

\paragraph{Camera reconstruction.}%
\label{sec:camera_annot}

The camera parameters $P_i^j$ were reconstructed by using COLMAP~\cite{sfm}, an off-the-shelf Structure-from-Motion software, on a uniformly-sampled subset of $N_{\mathcal{V}_i}=200$ frames per video.
Video frames are rectified to account for non-linear distortion
and the camera parameters are given as $4 \times 4$ projection matrices.

COLMAP does not reconstruct cameras correctly for all videos.
We use a semi-automated method to assign a \emph{camera reconstruction quality score} to each video and consider only the best 2,966 videos for the benchmark (1,234 cat and 1,732 dog videos).
Even so, we release the full set of 4,200 videos as future SfM improvements will likely make it possible to use many more for reconstruction.
For camera scoring, we follow the CO3D approach and use a human-in-the-loop active learning scheme.
In each active learning iteration, a human looks at some videos with the generated COLMAP camera tracks and assigns a binary label indicating the subjective \textit{correctness} of the tracking.
An SVM classifier is then trained to predict camera correctness given a set of features characterizing the reconstruction.
The SVM is evaluated on all videos and the results are given back to the annotator, who then labels false positives and false negatives to improve the SVM's decision boundary.
In total, 1,000 videos of cats and dogs were manually labelled to train the SVM\@.

%% file: 004_background.tex
\section{Few-shot dynamic new-view synthesis}%
\label{sec:method}

We discuss first necessary background and then our reconstruction method.

\begin{figure*}[t!]
\centering
\includegraphics[width=0.9\textwidth]{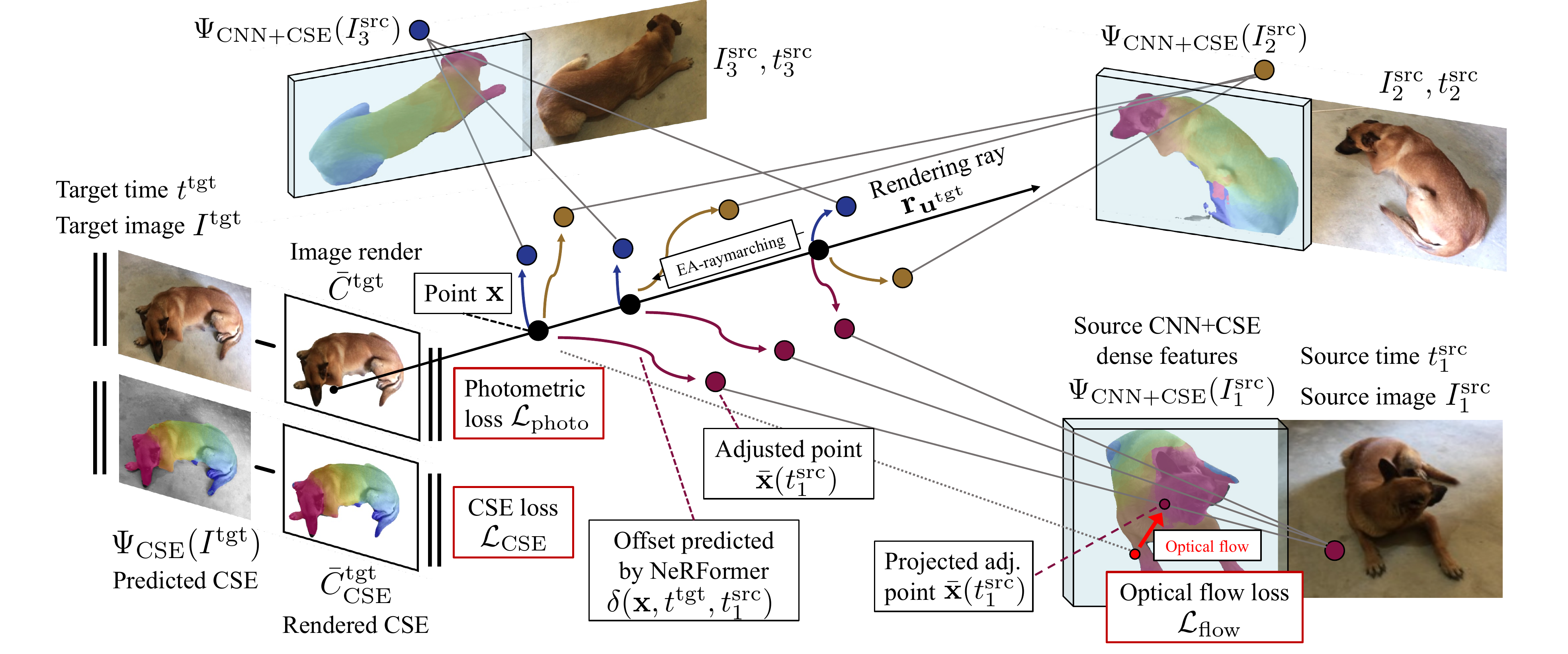}
\caption{\textbf{\themethod~reconstructs non-rigid 3D shape given few source views.}
It estimates the 3D motion of the shape by predicting source-view-specific offsets for each point on a projection ray emitted from the target image.
The offset points are then projected to each source view and the image features and CSE descriptors~\cite{Neverova2020cse} are sampled.
\themethod~then aggregates the information for each 3D point and predicts the point-specific colour, opacity and surface embedding.
The latter is then rendered using Emission Absorption rendering.}%
\label{fig:method}
\end{figure*}

\subsection{Background}%
\label{sec:background}

Many recent new-view synthesis methods represent the 3D object or scene as a \emph{radiance field}, i.e., a function $\sigma = f^\sigma(\bx), \bc = f^\bc(\bx, \br)$ that maps 3D points $\bx \in \mathbb{R}^3$ and viewing directions $\br \in \mathbb{S}^2$ to a corresponding colour $\bc \in \mathbb{R}^3$ and opacity $\sigma \in [0, 1]$.
The radiance field is called \emph{neural} when the function $f$ is implemented by a neural network.
This network is often a multi-layer perceptron (MLP) applied to a spatial encoding of the 3D point $\bx$.
We use the \emph{harmonic encoder}
$
\gamma(a) = [a,\sin(a), \cos(a), \dots, \sin(2^\Gamma a), \cos(2^\Gamma a)] \in \mathbb{R}^{2(\Gamma+1)},
$
where $\Gamma \in \mathbb{N}$ is a hyperparameter.

The neural field $f$ can be either fitted to a single object~\cite{mildenhall20nerf:} or to a large number of objects of the same category, learning a prior on the shape and appearance of objects of that type~\cite{co3d,sitzmann19scene}.
We discuss both cases below.

\paragraph{Fitting a neural field to a single object.}

The field $f$ is fitted to images of a given object via \emph{differentiable rendering}.
A renderer produces from the field $f$ and a target camera $\Ptgt$ a corresponding RGB image $\barItgt \in \mathbb{R}^{3 \times H \times W}$.
The parameters of the function $f$ are then optimized with stochastic gradient descent to minimize the photometric reconstruction error $\mathcal{L}_\text{photo} = \| \Itgt - \barItgt \|^2$ and mask error $\mathcal{L}_\text{mask} = \mathcal{L}_\text{BCE}(\Mtgt,\barMtgt)$, where $\mathcal{L}_\text{BCE}$ is the binary cross entropy between the known and rendered masks.

Rendering uses the \emph{Emission-Absorption} (EA) model~\cite{henzler19escaping,mildenhall20nerf:}.
In order to render the color of a pixel $\bu \in \{1, \dots W\} \times \{1, \dots H\}$ in the target image, one considers the ray $\br_\bu \in \mathbb{S}^2$ from the camera center through the pixel.
A number $N_S$ of 3D points $( \bx_i )_{i=1}^{N_S}$ are sampled along the ray at equal distances $\Delta$ and input to the field to obtain their colors and opacities
$
(\bc_i, \sigma_i) = f(\bx_i, \br_\bu).
$
The color $\bar \bc_\bu \in \mathbb{R}^3$ of the pixel $\bu$ of $\barItgt$ is the weighted combination
$
\bar \bc_\bu = \sum_{i=1}^{N_S} w(\bx_i) \bc_i
$
of the sample colors.
The weights are given by
$
w(\bx_i) = T(\bx_i) (1 - e^{-\sigma_i \Delta})
$
where the \emph{transmission coefficient} is
$
T(\bx_i) = e^{-\sum_1^{i-1} \sigma_i \Delta}
$.

\paragraph{Learning a category-level prior using a neural field.}

Fitting a neural field to a small number of views (e.g., fewer than 20) is highly ambiguous and a good results can only be obtained by using a prior to compensate for the missing information.
One way to do so is to learn an \emph{autoencoder}, where the neural field $f(\bx, \br, \bz) = (\bc, \sigma)$ is a function of an additional latent code $\bz \in \mathbb{R}^{D^\bz}$ predicted by an \emph{encoder} function
$
\bz(
\{(\Isrc_i, P_i)\}_{i=1}^{N_\text{src}}
)
$
of the $N_\text{src}$ available source images.
The autoencoder is learned form a large dataset and captures implicitly the required prior.

Methods differ in the way the autoencoder is designed.
\emph{Warp-Conditioned Embedding} (WCE~\cite{henzler2021unsupervised}) uses an autoencoder inspired by the geometry of image formation.
Given a source image $\Isrc$, a WCE $\bz_\text{WCE}(\bx, \Isrc)$ pools information about the 3D point $\bx$ by looking at its 2D projection:
\begin{equation}\label{e:wce}
\bz_\text{WCE}(\bx, \Isrc)
=
\Psi_\text{CNN}(\Isrc)[\pi_{\Psrc}(\bx)],
\end{equation}
where $\Psi_\text{CNN}(\Isrc) \in \mathbb{R}^{D^\bz \times H \times W}$ is a CNN that extracts dense 2D features from the source image and $\pi_{\Psrc}(\bx)$ is the projection of the 3D point on the source camera $\Psrc$.
The code 
$
\bz^\star_\text{WCE}
(
\bx,
\{(\Isrc_i, P_i)\}_{i=1}^{N_\text{src}}
)
$
of multiple source images is the concatenation of the mean and standard deviation of the WCEs of the individual images.

\newcommand{\butgt}{\bu}
\newcommand{\bxrutgt}{\bx}
\newcommand{\bxru}{\bx}
\newcommand{\deltaru}{\delta}

NeRFormer~\cite{co3d} extends WCE by further processing codes with a transformer network~\cite{vaswani17attention}.
Given a pixel $\bu$, it forms a $N_S \times N_\text{src}$ grid of $D^\bz$-dimensional WCE tokens
$
Z^{\br_\bu}_\text{WCE}  = \left[
  \bz_\text{WCE}(\bxru_j, \Isrc_i)
  \right]
  \in \mathbb{R}^{D^\bz \times N_\text{S} \times N_\text{src}}
$,
spanning the $N_\text{S}$ ray samples $\bxru_j$ and the $N_\text{src}$ source views $\Isrc_i$.
The neural field is implemented by a network
$
f'_\text{TR}(Z^{\br_\bu}_\text{WCE}) = (\bc_j, \sigma_j)_{j=1}^{N_\text{S}}
$
that predicts colors and opacities for all ray samples $\bx_j$.
It does so by alternating transformer layers that attend to tokens along the ray and source view dimensions, respectively.

%% file: 005_method.tex
\subsection{\themethod}%
\label{sec:our_method}

\themethod (\themethodabbrev) extends NeRFormer~\cite{co3d} to dynamic objects by:
$(i)$ predicting the trajectory of each rendered 3D point in order to accommodate for the non-rigid deformation of the object, and
$(ii)$ leveraging the Canonical Surface Embedding~\cite{Neverova2020cse} to help establish correspondences between different videos and object instances (see \cref{fig:method}).

\paragraph{Time Warp-Conditioned Embedding (TWCE).}

Both NeRF-WCE and NeRFormer assume a rigid object with a fixed pose across source frames $I^\text{src}$.
Given a 3D point $\bxrutgt$ along the ray $\br_\butgt$, a \emph{Time Warp-Conditioned Embedding} (TWCE) extends the WCE $z_\text{WCE}(\bxrutgt, I^\text{src})$ to also encode the \emph{timestamps} $t^\text{tgt}$ and $t^\text{src}$ of the source and target images:
\begin{equation}\label{eq:twce}
z_\text{TWCE}(\bxrutgt, I^\text{src}) = [
    z_\text{WCE}(\bxrutgt, I^\text{src}),
    \gamma(t^\text{tgt}),
    \gamma(t^\text{src})],
\end{equation}
where $\gamma$ is the usual harmonic encoding.
Adding the timestamps allows the neural field to `sense' time and model dynamic objects.

\paragraph{Modelling non-rigid deformations.}

According to~\cref{e:wce}, the WCE component of \cref{eq:twce} pools the information about the 3D point $\bxrutgt$ from its re-projection $\pi_{\Psrc}(\bxrutgt)$ on the source view $\Isrc$.
This equation fails to account for the fact that the physical 3D point can move and thus does not always lead to a valid correspondence.
In \themethodabbrev, we propose to explicitly compensate for the point motion.
We do so by predicting the location of the 3D point $\bxrutgt$ at time $t^\src$ as:
\begin{equation}\label{eq:offset}
\bar \bxrutgt(t^\src)
=
\bxrutgt + \deltaru(\bxrutgt,t^\tgt, t^\src),
\end{equation}
where $\deltaru(\bxrutgt,t^\tgt, t^\src) \in \mathbb{R}^3$ is the displacement of $\bxrutgt$ from the target time $t^\tgt$ to the source time $t^\src$, also known as \emph{scene flow}.
The scene flow $\deltaru(\bxrutgt,t^\tgt, t^\src)$ is predicted by a modified NeRFormer network
\begin{equation}\label{eq:flow-predictor}
\left( \deltaru(\bxrutgt_j, t^\tgt, t^\src) \right)_{j=1}^{N_\text{S}}
=
\mathcal{D}_\text{TR}(Z_\text{TWCE}^{\br_{\butgt}}),
\end{equation}
which takes as input the grid of TWCE tokens
$
Z^{\br_{\butgt}}_\text{TWCE} =
\left[
\bz_\text{TWCE}(\bxrutgt_j, \Isrc_i)
\right]
\in
\mathbb{R}^{D^\bz \times N_\text{S} \times N_\text{src}}.
$

\paragraph{\themethod.}

The network $\mathcal{D}_\text{TR}$ still takes as input the TWCE pooled from potentially incorrect 2D locations; its goal is to estimate such errors and `resolve' them by computing the adjusted 3D points $\bar\bx(t^\src)$.
The adjusted points are then used in a vanilla NeRFormer network $f'_\text{TR}$ to compute colors and opacities.
This is done by evaluating the function
\begin{equation}\label{eq:tracknerformer}
(\bc_j^{\br_{\butgt}}, \sigma_j^{\br_{\butgt}})_{j=1}^{N_\text{S}}
=
f'_\text{TR}(\bar Z^{\br_{\butgt}}_\text{TWCE}),
\end{equation}
which takes as input the resampled tokens
$
\bar Z^{\br_{\butgt}}_\text{TWCE} =
\left[
\bz_\text{TWCE}(\bar \bxrutgt_j(t_i^\src), \Isrc_i)
\right]
\in
\mathbb{R}^{D^\bz \times N_\text{S} \times N_\text{src}}.
$
Intuitively, by using the offset 3D points $\bar \bxrutgt_j(t_i^\src)$, the TWCEs pool information from pixels in the source views that are in proper correspondence to the target 3D point.

Note that \themethodabbrev differs significantly from prior works that also model point trajectories~\cite{Li2021nsff,luo20consistent,videonerf,Park2021hypernerf}:
while the latter retrain the deformation model from scratch for every reconstructed scene, \themethodabbrev trains a single model on a large dataset of videos, and later predicts deformations in a feed-forward manner on new videos without retraining, using an auto-encoder.

\input{tables/overfitting}
\input{tables/category}

\input{tables/ablation}

\paragraph{Flow-consistent \themethodabbrev.}

Since learning the scene flow $f^{\deltaru} = \deltaru(\bxrutgt,t^\tgt, t^\src)$ is generally ill-posed, we further constrain it to match the 2D optical flow via the following loss:
\begin{align}
\mathcal{L}_\text{flow} &=
\sum_{\butgt \in M^\text{flow}_{\text{tgt}\rightarrow\text{src}}}
\sum_{\bx_j \in \br^\bu} 
w(\bxrutgt_j)
e(\bxrutgt_j, t^\src),
\\
e(\bxrutgt_j, t^\src)
&=
\|
\butgt
+
\mathcal{F}_{\tgt \rightarrow \src}[
\butgt
]
-
\pi_{P^{\src}}(\bar \bxrutgt_j(t^\src))
\|_\epsilon,
\end{align}
where $\mathcal{F}_{\text{tgt} \rightarrow \text{src}}$ denotes RAFT~\cite{teed20raft:} optical flow predictions, mapping pixels from the target frame $I^\text{tgt}$ to their corresponding 2D locations in the source frame $I^\text{src}$, and
$
\| \mathbf{a} \|_\epsilon = \epsilon(\sqrt{1 + \frac{\| \mathbf{a} \|^2}{\varepsilon^2}} - 1)
$
is the soft Huber norm with cut-off threshold $\epsilon=0.001$.

The loss is computed only for the pixels $\butgt \in M^\text{flow}_{\text{tgt}\rightarrow\text{src}}$ that belong to the foreground ($M^\text{tgt}[\butgt]=1$) \emph{and} for which the optical flow is reliable, in the sense that it passes the standard forward-backward flow consistency check~\cite{flow_fwd_bwd}.
Furthermore, the factor $w(\bxrutgt_j)$ restricts the loss to only the 3D points $\bxrutgt_j$ that approximately lie on the surface of the object and that are visible (and thus contribute to the optical flow).
Finally, we stop gradient backpropagation from $\mathcal{L}_\text{flow}$ to $w(\bxrutgt_j)$ as we observed this to improve convergence.

\begin{figure*}[ht!]
\centering
\includegraphics[width=\textwidth]{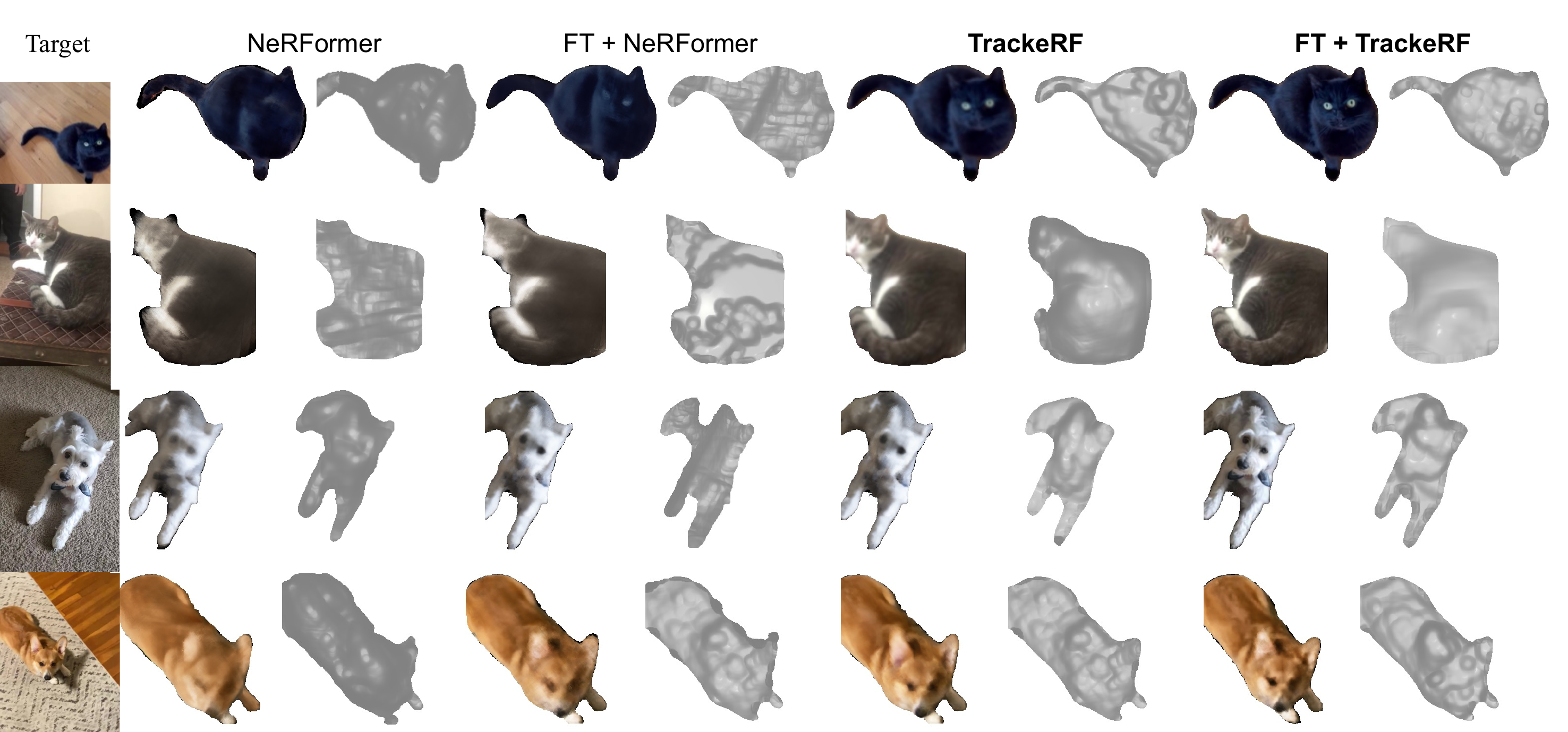}
\caption{\textbf{Many-Shot Single-Scene Reconstruction} (MSSSR) on \thedataset.
The prefix \method{FT+} denotes models pretrained on the entire \texttt{train-seen} subset of \thedataset for the FSCR task, and then fine-tuned to a novel \texttt{test} sequence following the MSSSR protocol.
\themethodabbrev results in smoother spatio-temporal interpolations and sharper and more accurate  new-view synthesis than baselines, and pre-training further improves the results.}%
\label{fig:single-seq-qual}
\end{figure*}

\begin{figure*}[ht!]
\centering
\includegraphics[width=\textwidth]{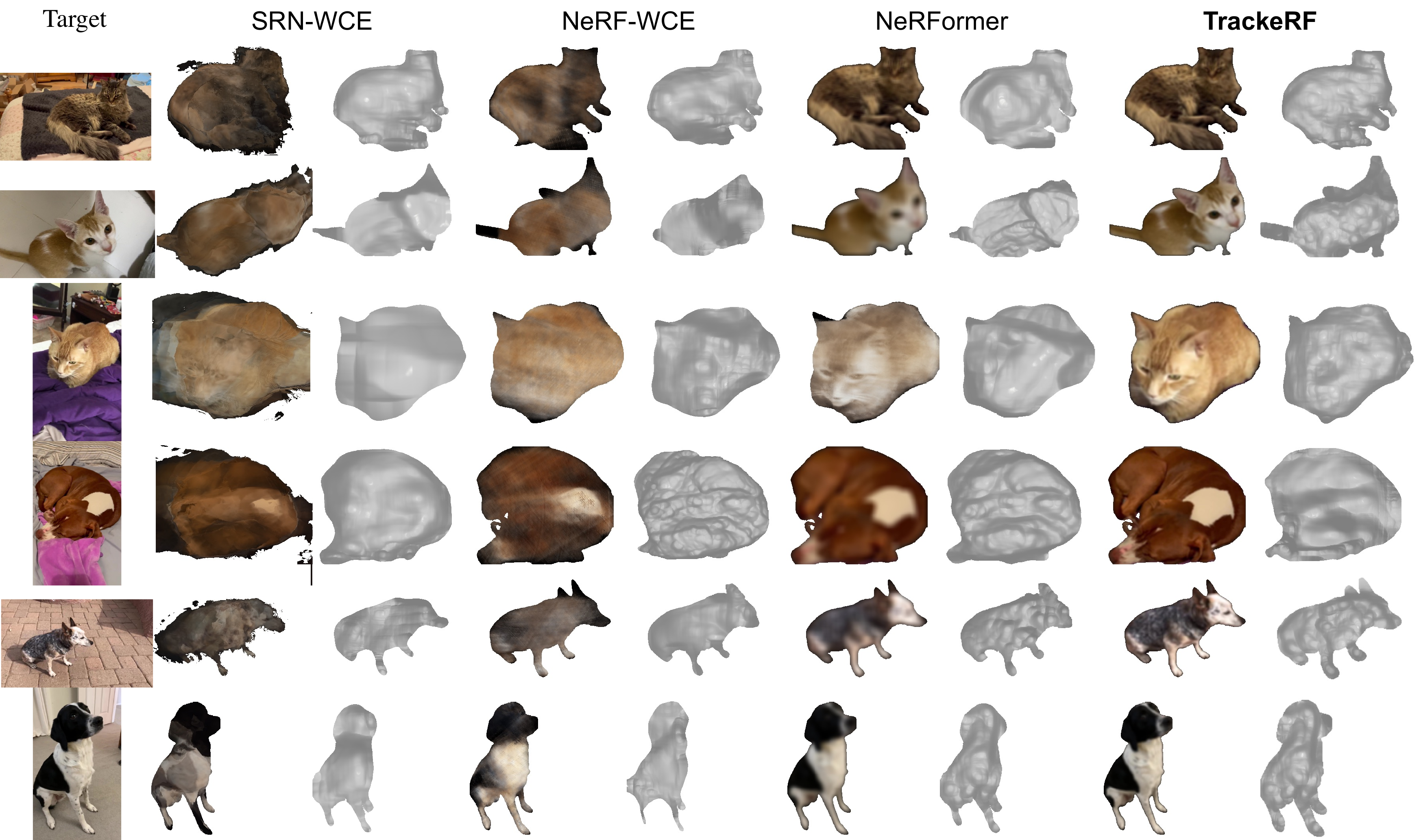}
\caption{\textbf{Few-Shot Category Reconstruction} (FSCR) on \thedataset.
\method{\themethodabbrev} results in sharper and more accurate new-view synthesis by compensating for the motion of the underlying 3D points in the source views.}%
\label{fig:multi-seq-qual}
\end{figure*}

\paragraph{Canonical Surface Embeddings.}

We can further help the model by incorporating category-specific information captured by the Canonical Surface Embedding (CSE) of~\cite{Neverova2020cse}.
A CSE assigns to each pixel $\bu$ of the object an embedding vector that uniquely identifies the corresponding point on the surface of the object (also known as a canonical map).
We use CSE models pretrained for cats and dogs, respectively.

We incorporate CSE in our formulation both as features and as a regularizer.
First, we concatenate the CSE vectors to the \emph{source view} embeddings already computed by the CNN, resulting in a function $\Psi_\text{CNN+CSE}: \mathbb{R}^{3 \times H \times W} \mapsto \mathbb{R}^{(D^\bz + D^\text{CSE}) \times H \times W}$.
This allows \themethodabbrev to more easily sense 2D correspondences, which helps it to reconstruct the scene flow vectors $\delta$.

Second, similar to~\cite{Yang2021banmo}, we task the model with predicting the canonical surface embeddings for the \emph{target view} by considering an extended field $f^\text{CSE}(\bx, \br, \bz) = (\bc, \sigma, \bC)$ which assigns to each 3D point $\bx$ a CSE vector $\bC \in \mathbb{R}^{D^\text{CSE}}$ in addition to a color $\bc$ and an opacity $\sigma$.
This field is supervised by rendering: in addition to the RGB colors, we also render the corresponding CSE vectors, and minimize the CSE rendering loss $\mathcal{L}_\text{CSE}$:
\begin{equation}
    \mathcal{L}_\text{CSE} 
    = 
    \|
    M^\text{tgt} \odot (\Psi_\text{CSE}(\Itgt) - 
    \bar \bC_\text{CSE}^\text{tgt})
    \|_\epsilon,
\end{equation}
where $\bar \bC_\text{CSE}^\text{tgt} \in \mathbb{R}^{D^\text{CSE} \times H \times W}$ are the rendered CSE embeddings generated with the same Emission-Absorption model used to render colors $\bar I^\text{tgt}$, and $\Psi_\text{CSE}(\Itgt) \in \mathbb{R}^{D^\text{CSE} \times H \times W}$ are  embeddings extracted from the target image $\Itgt$ by the pretrained CSE network $\Psi_\text{CSE}$.

\paragraph{Optimization.}

We minimize the total loss
$\lambda_\text{photo} \mathcal{L}_\text{photo} + \lambda_\text{flow} \mathcal{L}_\text{flow} + \lambda_\text{CSE} \mathcal{L}_\text{CSE}$
(with $\lambda_\text{photo}=1, \lambda_\text{flow}=1000, \lambda_\text{CSE}=10$) using Adam optimizer with an initial learning rate of $5 \times 10^{-4}$, which is decayed by a factor of 10 whenever the total loss plateaus.
During training, we render into a single known target view and randomly sample between 5 and 25 source views for each batch.

\begin{figure*}[h!]
\centering
\includegraphics[width=0.9\textwidth]{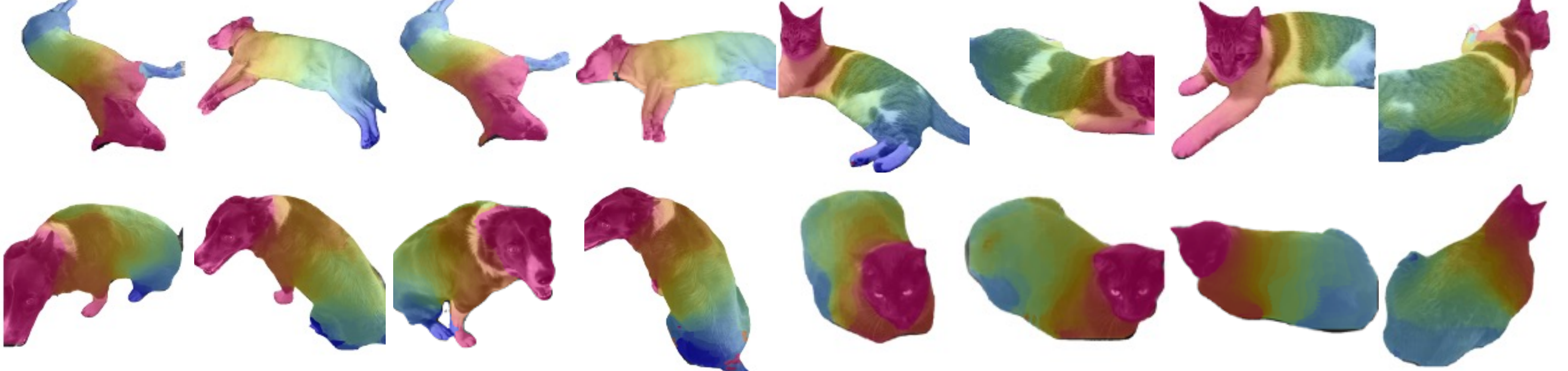}
\caption{\textbf{Visualization of the predicted Canonical Surface Embeddings rendered by \method{\themethodabbrev}.} \method{\themethodabbrev} learns dense correspondences which are consistent across different views of the animal.}%
\label{fig:dp-renders}
\end{figure*}

%% file: tables/overfitting.tex
\begin{table}
\newcommand{\ovrcolw}{0.27cm}%
\centering%
\small
\begin{tabular} {l @{\extracolsep{0.1cm}} c c c c }
$\downarrow$ Method                             & PSNR        & LPIPS   & IoU     & $\ell_1^\text{RGB}$  \\ \cmidrule{1-1} \cmidrule{2-5}  
        \method{SRN+AD} & 17.2  & 0.24 & 0.78 & 0.27 \\
        \method{SRN+TWCE} & 16.8 & 0.19 & 0.75 & 0.29 \\
        \method{TimeNeRF} & 17.3 & 0.18 & 0.72 & 0.20 \\
        \method{NeRF+TWCE} & 17.3 & 0.19 & 0.73 & 0.46 \\ 
        \method{NeRFormer+TWCE} & 18.6 & 0.17 & 0.82 & 0.21 \\ 
        \method{NSFF} & 20.2 & 0.17 & - & {\ul 0.19} \\
        \textbf{\method{\themethodabbrev} (ours)} & {\ul 21.4} & {\ul 0.15} & \textbf{0.91} & \textbf{0.17} \\
        \midrule
        \method{FT+NeRF+TWCE} & 17.7 & 0.19 & 0.82 & 0.30 \\
        \method{FT+NeRFormer+TWCE} & 20.5 & {\ul 0.15} & {\ul 0.88} & 0.20 \\
        \textbf{\method{FT+\themethodabbrev} (ours)} & \textbf{23.1} & \textbf{0.13} & \textbf{0.91} & \textbf{0.17} \\
        \bottomrule
    \end{tabular}
\caption{\textbf{Many-Shot Single-Scene Reconstruction} (MSSSR) on \thedatasetabbrev.
Models prefixed by the string \method{FT-} were pre-trained on the FSCR task and later separately fine-tuned to each scene of the MSSSR task
(\textbf{best}/\uline{2nd best}).
}%
\label{tab:overfitting_main}
\end{table}

%% file: tables/category.tex
\begin{table*}
\setlength\tabcolsep{0.05cm}
\small
\centering
\newcommand{\isad}{\textsuperscript{\textdagger}}
\begin{tabular}{
@{}
l@{\extracolsep{0.1cm}}
rrrr@{\extracolsep{0.1cm}}
rrrr@{\extracolsep{0.1cm}}
rrrrr@{\extracolsep{0.1cm}}
rrrrr@{\extracolsep{0.1cm}}
@{}
}
               & \multicolumn{8}{c}{\textbf{(a) Average statistics}}
               & \multicolumn{10}{c}{\textbf{(b) PSNR @ \# source views}} \\ \cmidrule{2-9} \cmidrule{10-19}
\multicolumn{1}{r}{Frame set $\rightarrow$}
& \multicolumn{4}{c}{\traintestset}
& \multicolumn{4}{c}{\testtestset}
& \multicolumn{5}{c}{\traintestset}
& \multicolumn{5}{c}{\testtestset}
\\
               $\downarrow$ Method                         & PSNR & LPIPS & $\ell_1^\text{RGB}$ & IoU  & PSNR & LPIPS & $\ell_1^\text{RGB}$  & IoU       &  25 & 20 & 15 & 10 & 5 & 25  & 20 & 15 & 10  & 5 \\
                \cmidrule{1-1} \cmidrule{2-5} \cmidrule{6-9} \cmidrule{10-14} \cmidrule{15-19} 


\textbf{\method{\themethodabbrev} (ours)} & \textbf{19.1} & \textbf{0.16} & \textbf{0.33} & \textbf{0.80} & \textbf{19.6} & \textbf{0.17} & \textbf{0.31} & \textbf{0.82} & \textbf{19.7} & \textbf{19.6} & \textbf{19.6} & \textbf{18.8} & \textbf{17.6} & \textbf{21.0} & \textbf{20.0} & \textbf{20.0} & \textbf{18.2} & \textbf{17.9} \\
\method{NeRFormer+TWCE} \cite{reizenstein21common} & \uline{16.3} & 0.19 & \uline{0.39} & \uline{0.73} & \uline{16.6} & \uline{0.18} & \uline{0.36} & \uline{0.76} & \uline{16.5} & \uline{17.2} & \uline{16.6} & \uline{16.3} & \uline{14.9} & \uline{17.7} & \uline{16.9} & \uline{17.0} & \uline{15.6} & \uline{15.7} \\
\method{NeRF+TWCE} \cite{henzler2021unsupervised} & 14.6 & 0.20 & 0.48 & 0.60 & 14.2 & 0.20 & 0.44 & 0.60 & 14.8 & 15.1 & 14.6 & 14.6 & 14.0 & 15.9 & 15.3 & 15.0 & 14.4 & 15.0 \\
\method{SRN+TWCE} & 13.9 & \uline{0.18} & 0.52 & 0.53 & 14.2 & \uline{0.18} & 0.49 & 0.53 & 13.7 & 14.6 & 14.3 & 13.7 & 13.1 & 15.0 & 14.4 & 14.3 &13.3 & 14.2 \\
\method{SRN+AD} \cite{sitzmann19scene} & 15.5 & 0.19 & 0.40 & 0.66 & \multicolumn{4}{c}{-} & 15.0 & 16.2 & 15.5 & 15.1 & \multicolumn{5}{c}{-}  \\ \bottomrule
\end{tabular}
\caption{\textbf{Few-Shot Category Reconstruction (FSCR) results on CoP3D.}
Metrics are averaged over both categories of \thedatasetabbrev (cats and dogs).
We report: (a) average results over the whole dataset, and (b) PSNR depending on the number of source views $N^\src$
(\textbf{best}/\uline{2nd best}).
}%
\label{tab:category_main}
\end{table*}


%% file: tables/ablation.tex
\begin{table}[t]
\footnotesize%
\centering%
    \begin{tabular}{ccc|cccc}\toprule
    predict $\delta$  & $\mathcal{L}_\text{flow}$ & $\mathcal{L}_\text{CSE}$ & PSNR & LPIPS & $\ell_1^{\text{RGB}}$ & IoU  \\ \midrule
    \boldxmark & \boldxmark & \boldxmark & 16.6 & 0.18 & 0.36 & 0.76 \\
    \boldcheckmark        &    \boldxmark        & \boldxmark & 17.3 &	0.18 & 0.35  & 0.77 \\
            \boldxmark     &      \boldxmark     &        \boldcheckmark                     & 17.6 &	0.18 &	0.35 & 0.80 \\
    \boldcheckmark     &     \boldxmark      & \boldcheckmark              &  17.8 &	0.18 &	0.35 & 0.77 \\
    \boldcheckmark            &    \boldcheckmark        &           \boldxmark                & 18.9 &	\textbf{0.17} &	0.32 & 0.80 \\
    \boldcheckmark            &      \boldcheckmark      & \boldcheckmark              & \textbf{19.6} &	\textbf{0.17} &	\textbf{0.31} & \textbf{0.82} \\
    \bottomrule
    \end{tabular}
    \caption{Ablation study of \themethod~ on the \texttt{test-unseen} subset of \thedatasetabbrev few-view reconstruction (\textbf{best} in bold).
    \label{tab:ablation_tab}
    }
\end{table}

%% file: 006_exps.tex
\section{Experiments}%
\label{sec:exps}

\paragraph{Data.}

For evaluation, similar to CO3D~\cite{reizenstein21common}, the frames in the videos are split into four sets, as follows.
50 videos in each category (cats and dogs) are selected at random as \texttt{test} videos and the others are used as \texttt{training} videos.
Frames in each video (training and test) are further split by considering contiguous blocks of 15 frames as \texttt{known}, interleaved by blocks of 5 frames as \texttt{unseen}.
The known frames are used as input to reconstruction algorithms whereas the unseen frames are only used for evaluation, providing the unseen camera parameters and timestamps, but withholding the images and masks, which must be predicted.

\paragraph{Tasks.}

We defined two benchmark tasks, also similar to CO3D.
The first task, \emph{Many-Shots Single-Scene Reconstruction} (MSSSR), is analogous to the setup explored in new-view synthesis approaches like NeRF~\cite{mildenhall20nerf:}.
The goal is to reconstruct the \texttt{unseen} frames from all the \texttt{known} frames in each of 10 \texttt{test} videos for each category (cats and dogs).
We consider only 10 videos for evaluation because in the MSSSR setting a separate model is usually `overfitted' to each video, which is expensive.

The other task \emph{Few-Shots Category Reconstruction} (FSCR) is instead aimed at learning a category-specific prior and using it for reconstructing objects from a small number of views.
In the training phase, the model is given access to the \texttt{train-seen} subset of video frames (including masks, cameras and timestamps).
In the testing phase, the model receives $N_{\src} \in \{5,10,15,20,25\}$ \texttt{known} source video frames
$
\{(\Isrc_i, M^\src_i, \Psrc_i, t^\src_i)\}_{i=1}^{N^\src}
$
from a test video.
It also receives the camera parameters $\Ptgt$ and timestamp $t^\tgt$ of an \texttt{unseen} target frame and is tasked with predicting the target image $\barItgt$ and mask $\bar M^\tgt$.

\paragraph{Evaluation.}

We test the quality of reconstructed unseen images using several metrics:
the \emph{Peak Signal-to-Noise Ratio} (PSNR),
the $\ell_1$ distance ($\ell_1^\text{RGB}$),
the \emph{Learned Perceptual Image Patch Similarity} (LPIPS~\cite{lpips}) between the predicted and ground-truth images, and the
\emph{Intersection-over-Union} (IoU) between the predicted and ground-truth object masks.

\subsection{Many-Shots Single-Scene Reconstruction}

\paragraph*{Baselines.}

For the MSSSR task, we consider the following baselines:
Scene Representation Networks (\method{SRN})~\cite{sitzmann19scene},
\method{SRN} using the Time-Warp-Conditioned Embeddings (\method{SRN-TWCE})~\cite{sitzmann19scene,henzler2021unsupervised},
time-conditioned variant of Neural Radiance Fields~\cite{mildenhall20nerf:}, which extends NeRF by appending the harmonic encoding of a frame's timestamp to the spatial encoding of the 3D point coordinates (\method{Time-NeRF}),
NeRF with Time-Warp-Conditioned Embedding (\method{NeRF-TWCE}),
NeRFormer~\cite{co3d} using TWCE instead of WCE (\method{NeRFormer-TWCE}) and
Neural Scene Flow Fields (\method{NSFF})~\cite{Li2021nsff}.

\paragraph*{Results.}

Quantitative and qualitative results are given in \cref{tab:overfitting_main} and \cref{fig:single-seq-qual}, respectively.
It can be seen that \themethodabbrev outperforms all alternatives in all quantitaive metrics, and 
the qualitative renders look visually superior and less blurry compared to each of the baselines. 

\subsection{Few-Shots Category Reconstruction}

\paragraph*{Baselines.}

For the FSCR problem, we consider the same baselines as before but only if they can be adapted to this setting.
Specifically, we consider:
\method{NeRFormer-TWCE},
\method{NeRF-TWCE},
\method{SRN-TWCE},
and our \method{\themethodabbrev}.
We also consider \method{SRN+AD} from~\cite{reizenstein21common}, which conditions an \method{SRN} model on an additional latent code which is optimized as a scene-specific free parameter together with the rest of the network weights.

\paragraph*{Results.}

Quantitative and qualitative results are given in \cref{tab:category_main} and \cref{fig:multi-seq-qual}, respectively.
Once more, \themethodabbrev outperforms all other baseline methods, this time with a wider margin of improvements.
This is compatible with the fact that few-shot reconstruction depends more on the quality of the learned 3D prior.
Qualitatively,  \themethodabbrev produces significantly smoother spatial and temporal interpolations.

\paragraph*{Ablations.}

In \cref{tab:ablation_tab}, we evaluate the contribution of each of the \themethodabbrev's components for reconstructing dynamic objects, by switching them ``off'' and measuring the 
change in reconstruction accuracy in the FSCR setting.
The latter shows that there is a significant performance drop when any of the components is removed, which justifies the design choices made.
We further visualize the rendered continuous surface embeddings (CSEs) in from different viewpoints \cref{fig:dp-renders}.
This shows that \themethodabbrev can learn to interpolate the CSEs smoothly and learn correct dense correspondences across different views of the animals.

\subsection{Single-scenes with pre-training}

Finally, we test if models for MSSSR, which are `overfitted' to a single video at a time, can benefit from category-centric pretraining of FSCR\@.
Specifically, we follow the same evaluation procedure as in the MSSSR case, but the weights of each model are initialized by pre-training in the FSCR setting on the appropriate category.
This results in the following model variants:
\method{FT+NeRFormer-TWCE},
\method{FT+NeRF-TWCE},
and \method{FT+\themethodabbrev}.

The results in~\cref{tab:category_main} and~\cref{fig:multi-seq-qual} for the \method{FT+} models show that indeed pre-training improves results across the board, further demonstrating how \thedatasetabbrev can be used to learn category-specific 3D priors by simply fine-tuning the category-centric model on to the new video.

%% file: 007_conclusion.tex
\section{Conclusions}

We have introduced \thedataset, a new large-scale dataset to explore the problem of new-view synthesis of deformable objects.
The dataset consists of 4,200 videos of cats and dogs collected in the wild using smartphone devices.
Our new dataset supports research in 4D reconstruction from casually-recorded videos, which can be very impactful in applications such as VR \& AR\@.
The data allows to learn 4D reconstruction category-priors, that are useful for reconstructing non-rigid objects with better visual quality and from less data.
We have demonstrated this idea by introducing a new method, \themethod, which learns a prior on the deformation of objects in videos.
We have further demonstrated the benefit of these learned category priors by improving the quality of single-sequence reconstruction by pre-training the model for category-level reconstruction first, and then fine-tuning on the new sequence.

%% file: 008_appendix.tex
\setcounter{figure}{0} \renewcommand{\thefigure}{\Roman{figure}}
\setcounter{table}{0} \renewcommand{\thetable}{\Roman{table}}

\begin{strip}%
 \centering
 \Large
 \textbf{%
{
\Large Common Pets in 3D\@:\\
Dynamic New-View Synthesis of Real-Life Deformable Categories
}
\\
 \vspace{0.3cm} 
 \textit{Supplementary material}
 }\\
\vspace{0.5cm}
\includegraphics[width=\textwidth]{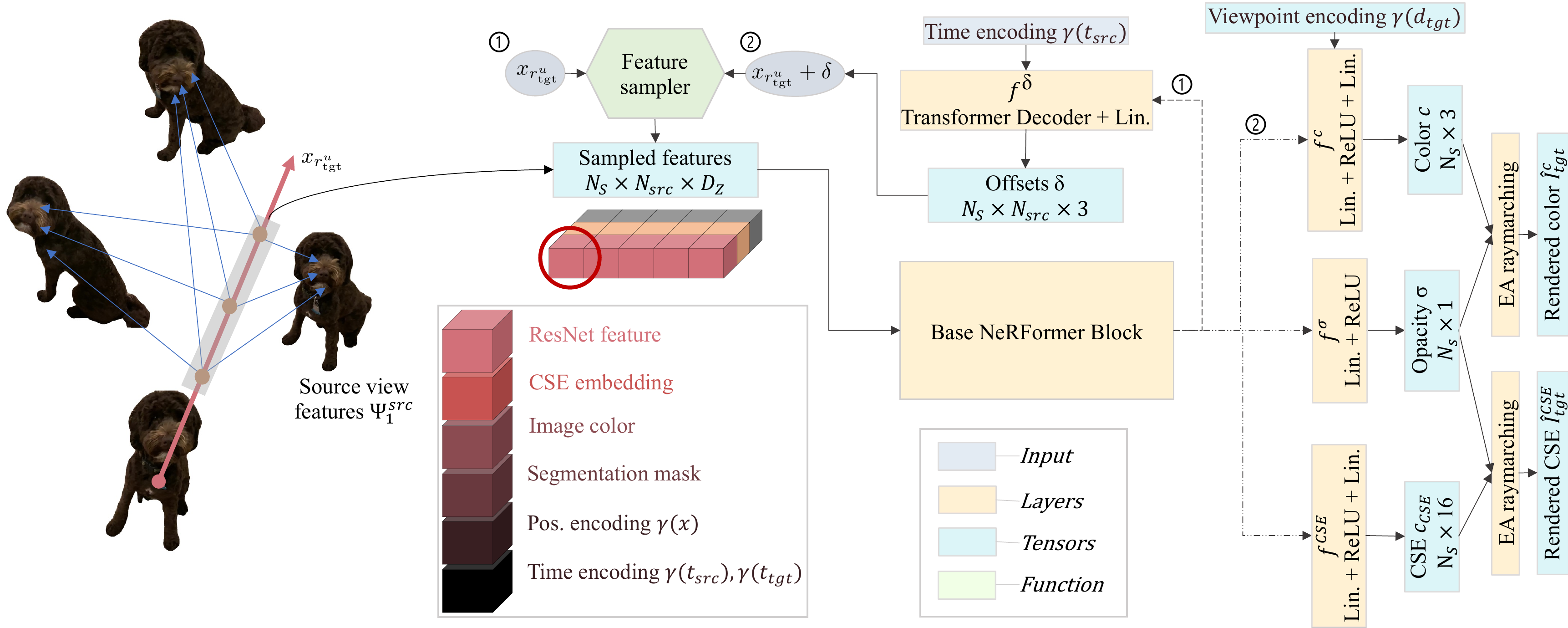}
\captionof{figure}{
\textbf{The architecture of \themethod.} 
First, source image features are bilinearly sampled given the points on the target ray $\br^\bu_\tgt$. 
The sampled features are then processed with the Base NeRFormer Block (with the same architecture as in \cite{reizenstein21common}) to generate intermediate features.
The latter enters the offset prediction head $\mathcal{D}_\text{TR}$ that generates per-point offsets $\delta$ (\Circled{1} in the figure).
The source images are then sampled again at the locations of projected offset-adjusted ray-points.
The resampled features enter Base NeRFormer Block again to predict a new intermediate feature grid which enters 3 final heads that predict colour, opacity, and CSE embedding for each ray point (\Circled{2} in the figure).
The final color and CSE render is formed by Emission-Absorption ray marching over the predicted opacities, colors, and CSE embeddings of the ray-points.
\label{fig:architecture}}
\end{strip}

\section{Network architecture}

The network architecture is summarized in \Cref{fig:architecture}.
The sampled grid of TWCE encodings 
$Z^{\br_{\butgt}}_\text{TWCE}$, or the adjusted tokens $\bar Z^{\br_{\butgt}}_\text{TWCE}$,
are processed with Base NeRFormer Block which has the same architecture as the vanilla NeRFormer from \cite{co3d}.
The output of the NeRFormer block is a set of intermediate features that are converted to either: 1) scene flow $\delta$ with the offset predictor $\mathcal{D}_\text{TR}$ during the first pass or, 2) are converted to colors $\bc$, opacities $\sigma$, or CSE embeddings $\bC$ with the final head $f'_\text{TR}$ during the second pass.
The offset predictor $\mathcal{D}_\text{TR}$ is implemented with a single-layer transformer decoder block that takes as input the intermediate features from the base NeRFormer block together with the encoding of the source time of each sampled point $\gamma(\tsrc)$, and outputs the per-point offsets.

\section{Mask annotations for \thedatasetabbrev}
\label{sec:mask_annot}

To generate the object masks $M_i^j$, we input each frame $I_i^j$ to the PointRend semantic segmentation network~\cite{pointrend}, extracting
$
N_i^j
$
candidate masks
$
\hat M_i^j \in [0, 1]^{N_i^j \times H \times W}.
$
To track a single foreground object across the video, we use a Hidden Markov Model to generate the set 
$\{M_i^t | M_i^t \in [0, 1]^{H \times W}\}_{j=1}^{N_{\mathcal{V}_i}}$
containing a single mask per frame, using intersection-over-union (IoU) of masks' bounding boxes as a pairwise potential.

%% file: main.bbl
\begin{thebibliography}{10}
\providecommand{\url}[1]{\texttt{#1}}
\providecommand{\urlprefix}{URL }
\providecommand{\doi}[1]{https://doi.org/#1}

\bibitem{ahmadyan2020objectron}
Ahmadyan, A., Zhang, L., Wei, J., Ablavatski, A., Grundmann, M.: Objectron: A
  large scale dataset of object-centric videos in the wild with pose
  annotations. Proc. {CVPR}  (2021)

\bibitem{chen21snarf:}
Chen, X., Zheng, Y., Black, M.J., Hilliges, O., Geiger, A.: {SNARF:}
  differentiable forward skinning for animating non-rigid neural implicit
  shapes. arXiv.cs  \textbf{abs/2104.03953} (2021)

\bibitem{Choi2016tanks}
Choi, S., Zhou, Q.Y., Miller, S., Koltun, V.: {A Large Dataset of Object Scans}
   (2016), \url{http://arxiv.org/abs/1602.02481}

\bibitem{nerfflow}
Du, Y., Zhang, Y., Yu, H.X., Tenenbaum, J.B., Wu, J.: Neural radiance flow for
  4d view synthesis and video processing. In: 2021 IEEE/CVF International
  Conference on Computer Vision (ICCV). pp. 14304--14314. IEEE Computer Society
  (2021)

\bibitem{plenoxels}
Fridovich-Keil, S., Yu, A., Tancik, M., Chen, Q., Recht, B., Kanazawa, A.:
  Plenoxels: Radiance fields without neural networks. In: Proceedings of the
  IEEE/CVF Conference on Computer Vision and Pattern Recognition. pp.
  5501--5510 (2022)

\bibitem{ignition2020google}
GoogleResearch: Google scanned objects (September),
  \url{https://fuel.ignitionrobotics.org/1.0/GoogleResearch/fuel/collections/GoogleScannedObjects}

\bibitem{henzler19escaping}
Henzler, P., Mitra, N.J., Ritschel, T.: Escaping plato's cave using adversarial
  training: {3D} shape from unstructured {2D} image collections. In: Proc.
  {ICCV} (2019)

\bibitem{wce}
Henzler, P., Reizenstein, J., Labatut, P., Shapovalov, R., Ritschel, T.,
  Vedaldi, A., Novotny, D.: Unsupervised learning of 3d object categories from
  videos in the wild. In: Proceedings of the IEEE/CVF Conference on Computer
  Vision and Pattern Recognition. pp. 4700--4709 (2021)

\bibitem{henzler2021unsupervised}
Henzler, P., Reizenstein, J., Labatut, P., Shapovalov, R., Ritschel, T.,
  Vedaldi, A., Novotny, D.: Unsupervised learning of 3d object categories from
  videos in the wild. Proc. {CVPR}  (2021)

\bibitem{codenerf}
Jang, W., Agapito, L.: Codenerf: Disentangled neural radiance fields for object
  categories. In: Proceedings of the IEEE/CVF International Conference on
  Computer Vision. pp. 12949--12958 (2021)

\bibitem{flow_fwd_bwd}
Jeong, J., Lin, J.M., Porikli, F., Kwak, N.: Imposing consistency for optical
  flow estimation. In: Proceedings of the IEEE/CVF Conference on Computer
  Vision and Pattern Recognition. pp. 3181--3191 (2022)

\bibitem{neuman}
Jiang, W., Yi, K.M., Samei, G., Tuzel, O., Ranjan, A.: Neuman: Neural human
  radiance field from a single video. arXiv preprint arXiv:2203.12575  (2022)

\bibitem{kanazawa18learning}
Kanazawa, A., Tulsiani, S., Efros, A.A., Malik, J.: Learning category-specific
  mesh reconstruction from image collections. In: Proc. {ECCV} (2018)

\bibitem{kar2017learning}
Kar, A., H{\"a}ne, C., Malik, J.: Learning a multi-view stereo machine.
  Advances in neural information processing systems  \textbf{30} (2017)

\bibitem{pointrend}
Kirillov, A., Wu, Y., He, K., Girshick, R.: Pointrend: Image segmentation as
  rendering. In: Proceedings of the IEEE/CVF conference on computer vision and
  pattern recognition. pp. 9799--9808 (2020)

\bibitem{kokkinos2021learning}
Kokkinos, F., Kokkinos, I.: {Learning monocular 3D reconstruction of
  articulated categories from motion}. In: IEEE Conference on Computer Vision
  and Pattern Recognition (2021). \doi{10.1109/cvpr46437.2021.00178},
  \url{http://arxiv.org/abs/2103.16352}

\bibitem{viewformer}
Kulh{\'a}nek, J., Derner, E., Sattler, T., Babu{\v{s}}ka, R.: Viewformer:
  Nerf-free neural rendering from few images using transformers. arXiv preprint
  arXiv:2203.10157  (2022)

\bibitem{UMR}
Li, X., Liu, S., Kim, K., Mello, S.D., Jampani, V., Yang, M.H., Kautz, J.:
  Self-supervised single-view 3d reconstruction via semantic consistency. In:
  European Conference on Computer Vision. pp. 677--693. Springer (2020)

\bibitem{li20self-supervised}
Li, X., Liu, S., Kim, K., Mello, S.D., Jampani, V., Yang, M., Kautz, J.:
  Self-supervised single-view {3D} reconstruction via semantic consistency. In:
  Proc. {ECCV} (2020)

\bibitem{Li2021nsff}
Li, Z., Yu, F., Zollh{\"{o}}fer, M., Rhodin, H.: {Neural Scene Flow Fields for
  Space-Time View Synthesis of Dynamic Scenes}. In: IEEE Computer Society
  Conference on Computer Vision and Pattern Recognition (2021),
  \url{https://arxiv.org/abs/2011.13084}

\bibitem{Liu2020nsvf}
Liu, L., Gu, J., Lin, K.Z., Chua, T.S., Theobalt, C.: {Neural sparse voxel
  fields}. NeurIPS  (2020)

\bibitem{Liu2021neuralactor}
Liu, L., Habermann, M., Rudnev, V., Sarkar, K., Gu, J., Theobalt, C.: {Neural
  actor: Neural Free-view Synthesis of Human Actors with Pose Control}. ACM
  Transactions on Graphics  \textbf{40}(6),  1--16 (2021).
  \doi{10.1145/3478513.3480528}

\bibitem{Lombardi2019nv}
Lombardi, S., Simon, T., Saragih, J., Schwartz, G., Lehrmann, A., Sheikh, Y.:
  {Neural volumes: Learning dynamic renderable volumes from images}. ACM
  Transactions on Graphics  \textbf{38}(4) (2019).
  \doi{10.1145/3306346.3323020}

\bibitem{neuralvolumes}
Lombardi, S., Simon, T., Saragih, J., Schwartz, G., Lehrmann, A., Sheikh, Y.:
  Neural volumes: Learning dynamic renderable volumes from images. arXiv
  preprint arXiv:1906.07751  (2019)

\bibitem{luo20consistent}
Luo, X., Huang, J., Szeliski, R., Matzen, K., Kopf, J.: Consistent video depth
  estimation. {ACM} Trans. Graph.  \textbf{39}(4), ~71 (2020)

\bibitem{Mildenhall2020}
Mildenhall, B., Srinivasan, P.P., Tancik, M., Barron, J.T., Ramamoorthi, R.,
  Ng, R.: {NeRF: Representing Scenes as Neural Radiance Fields for View
  Synthesis}. In: European Conference on Computer Vision (2020),
  \url{http://arxiv.org/abs/2003.08934}

\bibitem{mildenhall20nerf:}
Mildenhall, B., Srinivasan, P.P., Tancik, M., Barron, J.T., Ramamoorthi, R.,
  Ng, R.: {NeRF}: Representing scenes as neural radiance fields for view
  synthesis. In: Proc. {ECCV} (2020)

\bibitem{autorf}
M{\"u}ller, N., Simonelli, A., Porzi, L., Bul{\`o}, S.R., Nie{\ss}ner, M.,
  Kontschieder, P.: Autorf: Learning 3d object radiance fields from single view
  observations. In: Proceedings of the IEEE/CVF Conference on Computer Vision
  and Pattern Recognition. pp. 3971--3980 (2022)

\bibitem{ngp}
M{\"u}ller, T., Evans, A., Schied, C., Keller, A.: Instant neural graphics
  primitives with a multiresolution hash encoding. arXiv preprint
  arXiv:2201.05989  (2022)

\bibitem{Neverova2020cse}
Neverova, N., Novotny, D., Khalidov, V., Szafraniec, M., Labatut, P., Vedaldi,
  A.: {Continuous Surface Embeddings}. In: NeurIPS (2020),
  \url{https://arxiv.org/abs/2011.12438}

\bibitem{niemeyer19occupancy}
Niemeyer, M., Mescheder, L.M., Oechsle, M., Geiger, A.: Occupancy flow: 4d
  reconstruction by learning particle dynamics. In: Proc. {ICCV} (2019)

\bibitem{park20deformable}
Park, K., Sinha, U., Barron, J.T., Bouaziz, S., Goldman, D.B., Seitz, S.M.,
  Martin{-}Brualla, R.: Deformable neural radiance fields. CoRR
  \textbf{abs/2011.12948} (2020)

\bibitem{Park2021hypernerf}
Park, K., Sinha, U., Hedman, P., Barron, J.T., Bouaziz, S., Goldman, D.B.,
  Martin-Brualla, R., Seitz, S.M.: {HyperNeRF: A Higher-Dimensional
  Representation for Topologically Varying Neural Radiance Fields}. ACM
  Transactions on Graphics  \textbf{40}(6) (2021).
  \doi{10.1145/3478513.3480487}, \url{http://arxiv.org/abs/2106.13228}

\bibitem{neuralparts}
Paschalidou, D., Katharopoulos, A., Geiger, A., Fidler, S.: Neural parts:
  Learning expressive 3d shape abstractions with invertible neural networks.
  In: Proceedings of the IEEE/CVF Conference on Computer Vision and Pattern
  Recognition. pp. 3204--3215 (2021)

\bibitem{peng20neural}
Peng, S., Zhang, Y., Xu, Y., Wang, Q., Shuai, Q., Bao, H., Zhou, X.: Neural
  body: Implicit neural representations with structured latent codes for novel
  view synthesis of dynamic humans. CoRR  \textbf{abs/2012.15838} (2020)

\bibitem{Pumarola2020}
Pumarola, A., Corona, E., Pons-Moll, G., Moreno-Noguer, F.: {D-NeRF: Neural
  radiance fields for dynamic scenes}. arXiv  (2020)

\bibitem{reizenstein21common}
Reizenstein, J., Shapovalov, R., Henzler, P., Sbordone, L., Labatut, P.,
  Novotny, D.: {Common Objects in 3D}: Large-scale learning and evaluation of
  real-life 3d category reconstruction. In: arXiv (2021)

\bibitem{co3d}
Reizenstein, J., Shapovalov, R., Henzler, P., Sbordone, L., Labatut, P.,
  Novotny, D.: Common objects in 3d: Large-scale learning and evaluation of
  real-life 3d category reconstruction. In: Proceedings of the IEEE/CVF
  International Conference on Computer Vision. pp. 10901--10911 (2021)

\bibitem{rematas21sharf:}
Rematas, K., Martin{-}Brualla, R., Ferrari, V.: {ShaRF}: Shape-conditioned
  radiance fields from a single view. arXiv.cs  \textbf{abs/2102.08860} (2021)

\bibitem{sfm}
Schonberger, J.L., Frahm, J.M.: Structure-from-motion revisited. In:
  Proceedings of the IEEE conference on computer vision and pattern
  recognition. pp. 4104--4113 (2016)

\bibitem{schoenberger2016sfm}
Sch\"{o}nberger, J.L., Frahm, J.M.: Structure-from-motion revisited. In: Proc.
  {CVPR} (2016)

\bibitem{sitzmann19scene}
Sitzmann, V., Zollh{\"{o}}fer, M., Wetzstein, G.: Scene representation
  networks: Continuous 3d-structure-aware neural scene representations. Proc.
  {NeurIPS}  (2019)

\bibitem{anerf}
Su, S.Y., Yu, F., Zollh{\"o}fer, M., Rhodin, H.: A-nerf: Articulated neural
  radiance fields for learning human shape, appearance, and pose. Advances in
  Neural Information Processing Systems  \textbf{34},  12278--12291 (2021)

\bibitem{blocknerf}
Tancik, M., Casser, V., Yan, X., Pradhan, S., Mildenhall, B., Srinivasan, P.P.,
  Barron, J.T., Kretzschmar, H.: Block-nerf: Scalable large scene neural view
  synthesis. arXiv preprint arXiv:2202.05263  (2022)

\bibitem{teed20raft:}
Teed, Z., Deng, J.: {RAFT:} recurrent all-pairs field transforms for optical
  flow. In: Proc. {ECCV} (2020)

\bibitem{Tretschk2020}
Tretschk, E., Tewari, A., Golyanik, V., Zollh{\"{o}}fer, M., Lassner, C.,
  Theobalt, C.: {Non-rigid neural radiance fields: Reconstruction and novel
  view synthesis of a deforming scene from monocular video}. arXiv
  \textbf{1}(1) (2020)

\bibitem{nrnerf}
Tretschk, E., Tewari, A., Golyanik, V., Zollh{\"o}fer, M., Lassner, C.,
  Theobalt, C.: Non-rigid neural radiance fields: Reconstruction and novel view
  synthesis of a dynamic scene from monocular video. In: Proceedings of the
  IEEE/CVF International Conference on Computer Vision. pp. 12959--12970 (2021)

\bibitem{Tulsiani2020}
Tulsiani, S., Kulkarni, N., Gupta, A.: {Implicit Mesh Reconstruction from
  Unannotated Image Collections}  (2020), \url{http://arxiv.org/abs/2007.08504}

\bibitem{vaswani17attention}
Vaswani, A., Shazeer, N., Parmar, N., Uszkoreit, J., Jones, L., Gomez, A.N.,
  Kaiser, L., Polosukhin, I.: Attention is all you need. In: NIPS (2017)

\bibitem{vid2actor}
Weng, C.Y., Curless, B., Kemelmacher-Shlizerman, I.: Vid2actor: Free-viewpoint
  animatable person synthesis from video in the wild. arXiv preprint
  arXiv:2012.12884  (2020)

\bibitem{Weng2022humannerf}
Weng, C.Y., Curless, B., Srinivasan, P.P., Barron, J.T.,
  Kemelmacher-Shlizerman, I.: {HumanNeRF: Free-viewpoint Rendering of Moving
  People from Monocular Video} (2022), \url{http://arxiv.org/abs/2201.04127}

\bibitem{wu21dove:}
Wu, S., Jakab, T., Rupprecht, C., Vedaldi, A.: {DOVE}: Learning deformable {3D}
  objects by watching videos. In: arXiv (2021)

\bibitem{videonerf}
Xian, W., Huang, J.B., Kopf, J., Kim, C.: Space-time neural irradiance fields
  for free-viewpoint video. In: Proceedings of the IEEE/CVF Conference on
  Computer Vision and Pattern Recognition. pp. 9421--9431 (2021)

\bibitem{yang21lasr:}
Yang, G., Sun2, D., Jampani, V., Vlasic, D., Cole, F., Chang, H., Ramanan, D.,
  Freeman, W.T., Liu, C.: {LASR}: Learning articulated shape reconstruction
  from a monocular video. In: Proc. {CVPR} (2021)

\bibitem{Yang2021banmo}
Yang, G., Vo, M., Neverova, N., Ramanan, D., Vedaldi, A., Joo, H.: {BANMo:
  Building Animatable 3D Neural Models from Many Casual Videos}  (2021),
  \url{http://arxiv.org/abs/2112.12761}

\bibitem{yariv20multiview}
Yariv, L., Kasten, Y., Moran, D., Galun, M., Atzmon, M., Basri, R., Lipman, Y.:
  Multiview neural surface reconstruction by disentangling geometry and
  appearance. In: Proc. {NeurIPS} (2020)

\bibitem{yu20pixelnerf:}
Yu, A., Ye, V., Tancik, M., Kanazawa, A.: {pixelNeRF}: Neural radiance fields
  from one or few images. arXiv.cs  \textbf{abs/2012.02190} (2020)

\bibitem{Yu2020pixelnerf}
Yu, A., Ye, V., Tancik, M., Kanazawa, A.: {pixelNeRF: Neural Radiance Fields
  from One or Few Images}  (2020), \url{http://arxiv.org/abs/2012.02190}

\bibitem{lpips}
Zhang, R., Isola, P., Efros, A.A., Shechtman, E., Wang, O.: The unreasonable
  effectiveness of deep features as a perceptual metric. In: Proceedings of the
  IEEE conference on computer vision and pattern recognition. pp. 586--595
  (2018)

\end{thebibliography}
